\documentclass[pmlr,twocolumn,10pt]{jmlr} 

\usepackage{booktabs}
\usepackage{graphicx}

\usepackage{multirow}
\usepackage{color}
\usepackage{svg}
\usepackage{xcolor,colortbl}
\usepackage{todonotes}
\usepackage[T1]{fontenc}

\usepackage[group-separator={,}]{siunitx}

\definecolor{light-gray}{gray}{0.83}
\newcommand{\GREY}{\cellcolor{light-gray}\bf} 

\theorembodyfont{\upshape}
\theoremheaderfont{\scshape}
\theorempostheader{:}
\theoremsep{\newline}

\title[Scope and Arbitration in EEG Classification]{Scope and Arbitration in Machine Learning Clinical EEG Classification}

\author{%
\Name{Yixuan Zhu} \Email{yixuan2.zhu@live.uwe.ac.uk}\\
\addr University of the West of England, UK
\AND
\Name{Luke J. W. Canham} \Email{luke.canham@nbt.nhs.uk}\\
\addr  North Bristol NHS Trust, UK
\AND
\Name{David Western} \Email{david.western@uwe.ac.uk}\\
\addr University of the West of England, UK
}
\begin{document}

\maketitle

\begin{abstract}
A key task in clinical EEG interpretation is to classify a recording or session as normal or abnormal.
In machine learning approaches to this task, recordings are typically divided into shorter windows for practical reasons, and these windows inherit the label of their parent recording. 
We hypothesised that window labels derived in this manner can be misleading – for example, windows without evident abnormalities can be labelled `abnormal' – disrupting the learning process and degrading performance. 
We explored two separable approaches to mitigate this problem: increasing the window length and introducing a second-stage model to arbitrate between the window-specific predictions within a recording. 
Evaluating these methods on the Temple University Hospital Abnormal EEG Corpus, we significantly improved state-of-the-art average accuracy from 89.8 percent to 93.3 percent. 
This result defies previous estimates of the upper limit for performance on this dataset and represents a major step towards clinical translation of machine learning approaches to this problem.
\end{abstract}

\paragraph*{Data and Code Availability}
Our study includes electroencephalography (EEG) datasets collected from \url{https://isip.piconepress.com/projects/tuh\_eeg/}. Our code is shared on \url{https://github.com/zhuyixuan1997/EEGScopeAndArbitration}.

\section{Introduction}
\label{sec:intro}

\subsection{Background}
\label{sec:background}
Electroencephalography (EEG) recordings are used for the diagnosis and monitoring of a wide range of neurological conditions. Classification of EEG recordings as normal or abnormal is an essential task in their clinical interpretation. Substantial research has been conducted on the application of machine learning to this task \citep{schirrmeister2017deep,amin2019cognitive,banville2021uncovering,banville2022robust,gemein2020machine,muhammad2020eeg,wagh2020eeg,alhussein2019eeg,roy2019chrononet}. 

Recent work in this field largely makes use of the Temple University Hospital Abnormal EEG Corpus (TUAB) \citep{lopez2017automated} for training and evaluation. TUAB is a labelled subset of the Temple University Hospital EEG Corpus (TUEG) \citep{obeid2016temple}. 

Since the presentation of the Deep4 convolutional neural network in 2017 \citep{schirrmeister2017deep} there have been only modest improvements in the accuracy of machine learning approaches to this problem, as measured on TUAB: 
from 85.4 percent (Deep4) up to 89.8 percent \citep{muhammad2020eeg} -- see \tableref{tab:models} for further detail. 
\citet{gemein2020machine} proposed that there may be an upper limit of around 90 percent accuracy in this task, based on known values of inter-rater agreement between human experts in conventional clinical practice.

\begin{table}[htbp]
\floatconts
    {tab:models}
    {\caption{Summary of state-of-the-art performance metrics for different models applied to abnormal EEG classification}}%

    \resizebox{\linewidth}{!}
    {
    \begin{tabular}{|c|c|c|c|c|}
    \hline
    \GREY Model  &  \GREY Accuracy & \GREY Sensitivity & \GREY Specificity \\\hline
    1D-CNN (T5-O1 channel)\citep{yildirim2020deep} & 79.3 \% &71.4  \% &  86.0 \%   \\\hline
    1D-CNN (F4-C4 channel)\citep{yildirim2020deep} & 74.4 \% &55.6  \% & 90.7 \%   \\\hline
    Deep4 \citep{schirrmeister2017deep}     & 85.4 \% & 75.1 \%  & 94.1 \%  \\\hline
    TCN \citep{gemein2020machine} & 86.2 \% & &    \\\hline
    ChronoNet \citep{roy2019chrononet} & 86.6 \% & &    \\\hline
    Alexnet\citep{amin2019cognitive}  & 87.3 \% & 78.6 \%  & 94.7 \%  \\\hline
    VGG-16 \citep{amin2019cognitive}  & 86.6 \% & 77.8 \%  & 94.0 \%  \\\hline
    Fusion Alexnet\citep{alhussein2019eeg}	& 89.1 \%  & 80.2 \%&  96.7 \%  \\\hline
    \citep{muhammad2020eeg}	& 89.8 \% &81.3 \% & \bfseries{96.9} \%   \\\hline
    \bfseries{Proposed} & \bfseries{93.3} \%   &	\bfseries{92.0} \%   &	92.9 \%   \\\hline

    \end{tabular} 
}
 
\end{table}

A notable but little-discussed difference between conventional clinical practice and virtually all deep learning approaches is that in clinical practice, the label of normal/abnormal is applied to a full EEG session (i.e. a single clinical visit). 
In clinical practice, experts judge whether the patient exhibits abnormal brain activity based on all the recordings in the session, effectively resulting in a single label for that session. 
In most recent machine learning approaches, a typical full recording cannot be directly input into the model due to computational constraints -- a large input vector length necessitates a large number of parameters in the model. 
Instead, the recording is divided into smaller windows, with the added advantage of increasing the total number of examples available for training. 
For training purposes, each window inherits the label of its recording, while evaluation is typically performed on a per-recording basis by aggregating per-window outputs from the classifier.
We refer to this downstream aggregation as `arbitration'.

\citet{western2021automatic} noted that this inheritance of window/recording labels from broader session labels was potentially confounding to the machine learning process. For example, a session may be labelled as `abnormal' based on several temporally isolated abnormal graphoelements. Many windows in this session may be completely free of abnormal activity, yet they will carry `abnormal' labels in the training process. These labels are arguably false, depending on whether they are considered to apply to the signal within the window or to the wider session from which it is taken.

\subsection{Proposal}
\subsubsection{Overview}
An intuitive approach to address this problem of unrepresentative window labels is to expand the model to accept full (and perhaps multiple) recordings from a session, instead of individual windows, as the input. However, implementation of this approach is inhibited by notable practical challenges, including the reduced number of available samples (reducing the learning opportunity) and the increase in model complexity and associated computational expense. In the present study, we propose and evaluate two approaches to the problem: increasing the window length or introducing a second-stage machine learning model for arbitration. 

\subsubsection{Increasing the Window Length}

Increasing the window length can be considered to increase the accuracy of the labels applied to those windows. For example, it increases the probability of abnormal features being included within any given window from an abnormal session. 
It also increases the scope of the deep learning model, enabling it to use more information to make decisions.
Hence we established the following hypothesis.
However, it could also have negative effects. 
When the windows become longer, the number of windows available as training examples will decrease and the number of parameters in the model will increase, making the model more difficult to train.
Nonetheless, given the potential significance of temporally isolated abnormalities for the problem of clinical EEG classification, we hypothesised that the net effect of increased window length would be positive, as follows.

\begin{conjecture}[Increased Window Length]\label{con:WL}
Increasing the window length of a deep learning clinical EEG classifier will increase its accuracy and, more specifically, its sensitivity to abnormal cases.
\end{conjecture}

\subsubsection{Second-Stage Model for Arbitration}
\label{sec: pro_arb}

All recent state-of-the-art applications of deep learning to this problem make use of some form of windowing. 
Hence their final stage must be some form of arbitration to combine the outputs from each window into a single label for that recording, as depicted in \figureref{fig:EEG_classification}. 
One notable variation is the approach of \citet{alhussein2019eeg}, in which higher-dimensional features are fused across multiple windows in a multi-layer perceptron producing a per-recording classification, rather than applying arbitration to downstream class probabilities.
They demonstrate notable benefits from this approach, but it can be assumed that the required increase in model parameters introduces new challenges to the training process.
Otherwise, the arbitration stage receives little attention in prior literature in this area, implying that it has not been considered to be important to performance.
\citet{schirrmeister2017deep} explored the effects of total recording duration, but maintained a fixed window length of 60 s.
The arbitration function is not mentioned in that paper, but can be seen in the source code to be the mean of the per-window outputs (SoftMax-estimated probabilities of `normal' and `abnormal'). The per-recording class is then determined based on whichever mean probability is greater, `normal' or `abnormal' (argmax). 
This use of the `Mean' to arbitrate between multiple windows differs substantially from conventional clinical practice, in which the presence of any temporally isolated abnormality may be sufficient to warrant labelling the recording as abnormal if the rest of the recording appears normal.

Of course the confidence of a human expert or artificial intelligence in whether any given window contains an abnormality should never be absolute. 
It can instead be expressed as some fractional probability, $0<p<1$.
Hence, if class probabilities have been determined on a per-window basis then an arbitration algorithm must be used.
Two categories of clear-cut cases may exist:

\begin{description}
\item[clearly abnormal] one or more windows has a very high abnormality score ($p_{abnormal} > t_{upper}$)
\item[clearly normal] all windows have very low abnormality scores ($p_{abnormal} < t_{lower}$)
\end{description} 

Even in these cases, suitable decision thresholds $t_{lower}$ and $t_{upper}$ must be identified.
In all other cases, more complex decision-making is required.
However, given the dimensionality reduction achieved by the first stage model, we hypothesised that this arbitration model would not need to be as complex as the multilayer perceptron used by \citet{alhussein2019eeg} for fusion of higher-dimensional features across windows.
Furthermore, it could be trained separately to minimise the computational and practical expense of iterative development, while avoiding the need to reduce the number of training samples available to the first-stage deep learning model.
Thus thorough optimisation of the full model would be achieved more easily.

\begin{conjecture}[Arbitration]\label{con:arbitration}
For a deep learning clinical EEG classifier with windowing, the use of a shallow neural network as the arbitration stage will improve accuracy compared with conventional `Mean' arbitration or `No arbitration'.
\end{conjecture}

\begin{figure}[htbp]
\floatconts
  {fig:EEG_classification}
  {\caption{Generic diagram of a typical deep learning approach to clinical EEG classification, as used e.g. by \citet{schirrmeister2017deep}}}
  {\includegraphics[width=1\linewidth]{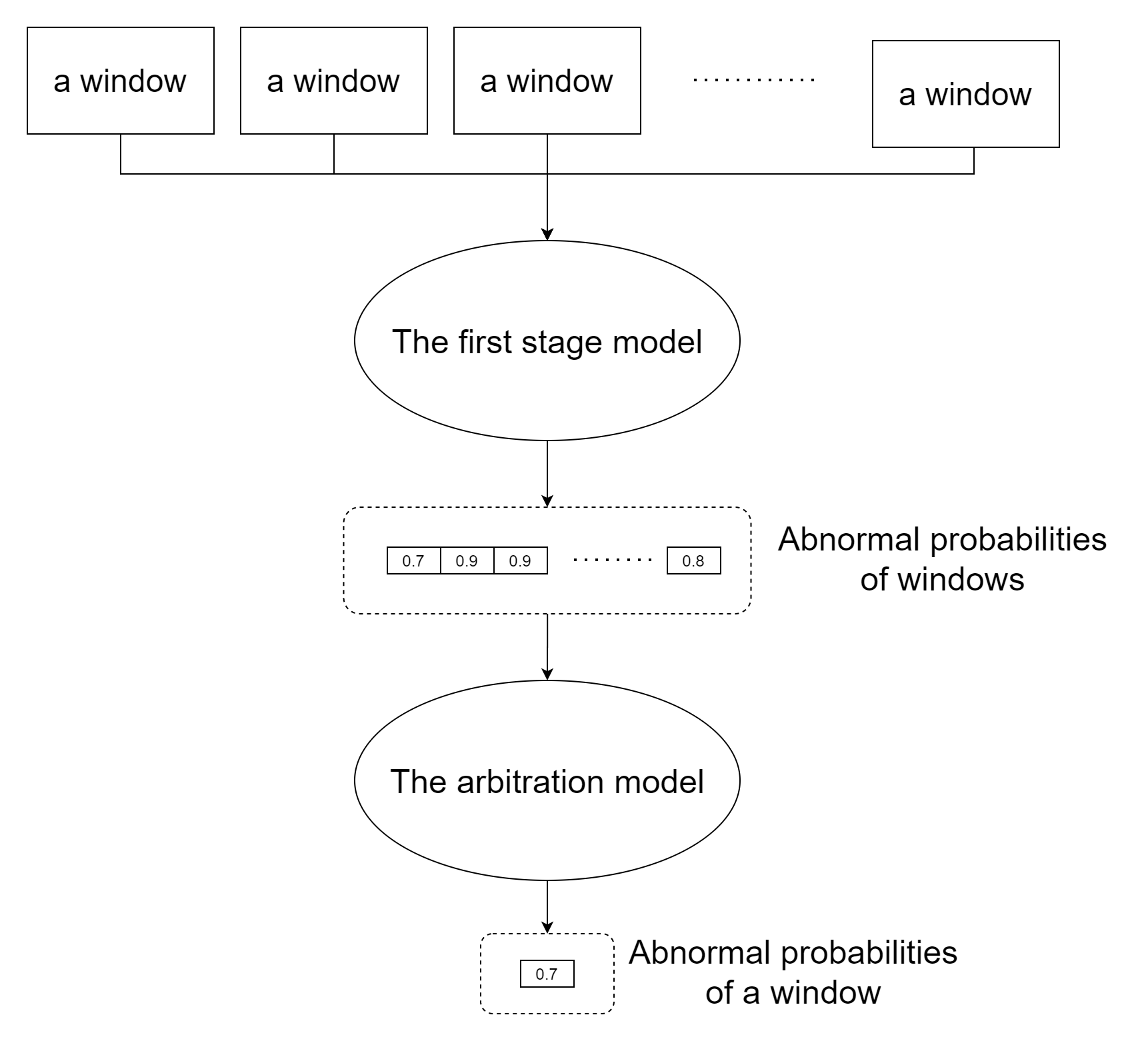}}
\end{figure}

\section{Method}
\subsection{Data}

TUEG \citep{obeid2016temple} is a rich archive of over 30,000 clinical EEG recordings collected at Temple University Hospital (TUH) from 2002 – present, using the standard 10-20 system of electrode placement. TUAB \citep{lopez2017automated} is a subset of 2993 recordings from TUEG that have been labelled as normal/abnormal and divided into training and evaluation sets. The training set contains 1371 normal sessions and 1346 abnormal sessions. The test set contains 150 normal sessions and 126 abnormal sessions. Also, only one file from each session was included in this corpus. All our current results are trained and tested on TUAB. 
Due to computational cost, all experiments related to window length are carried out on TUAB, including one-stage models, two-stage models and window length. TUAB has already been marked, so we use the original label and its original test-training split. To compare the results with other studies, we followed the pre-processing method in the Deep4 article, which TCN \citep{gemein2020machine} and Fusion Alexnet \citep{alhussein2019eeg} also used.

\subsection{First-Stage Model}

A recent study by \citet{gemein2020machine} demonstrated that the Temporal Convolutional Network (TCN) and Deep4 architectures offer near-state-of-the-art performance on TUAB, so we experimented with both of these as the first-stage model. 
Both are composed of blocks with convolutional and pooling layers. 
However, TCN replaces common convolution with dilated convolution and introduces a residual structure in the temporal block \citep{bai2018empirical}.
To achieve baseline performance consistent with past studies, we use the hyperparameters from \citet{schirrmeister2017deep} for Deep4 and from \citet{gemein2020machine} for TCN.

To explore whether our proposals are applicable to first-stage architectures other than convolutional networks, we also implemented a Vision Transformer (ViT) \citep{dosovitskiy2020image}.
For simplicity, the majority of our experiments focussed on Deep4 only.
We use Deep4 with the 60s, 180s, 300s, 400s, and 600s windows to perform reproducibility experiments on TUAB. For each Window length, five experiments were performed to avoid the influence of chance factors.

\subsection{Second-Stage Models for Arbitration}
As stated in \sectionref{sec: pro_arb}, the purpose of the arbitration stage is to combine the per-window class probabilities into a single classification of the EEG session. 
Previous work does not discuss arbitration, although some form of arbitration is inevitable where models are for trained on windows and evaluated on a per-recording basis (e.g. \citep{schirrmeister2017deep, gemein2020machine}). 
When looking through the code of Deep4 \citep{schirrmeister2017deep}, we found that they used the `Mean' method to integrate the results of windows. In some studies based on time-frequency images, they use a method, such as the Fourier transform\citep{alhussein2019eeg}, to freely choose the size of the image, thus eliminating the need for windowing.

Hence we employ `Mean' as a baseline arbitration model. As alternatives, we explore several implementations of multi-layer perceptrons as the arbitration model. These are distinguished from each other by the pre-processing of the input data (per-window scores) and the specific architecture used. The input pre-processing methods considered are as follows:

\begin{description}

\item[`Raw'] As shown in \figureref{fig:Raw and Histogram}, this approach inputs all the results in each recording directly(`Raw'). Since the number of windows in each recording is different, padding 0 at the end is required for less than 20 windows data. The value of 20 is chosen here to reflect the approach of \citet{schirrmeister2017deep,gemein2020machine}, which used 1-minute windows and a maximum of 20 minutes per recording.  

\item[`Histogram'] Being intended as a flexible approach to handling variations in recording length, a histogram (\figureref{fig:Raw and Histogram}) was calculated from the per-window scores. The range of potential per-window abnormality scores (0-1) was divided into ten equal bins. The input to the model was a vector of length ten containing, for each bin, the proportion of windows with scores in the corresponding range, as shown in Figure 6. 
Bin counts were divided by the total number of windows as a form of normalisation.

\item[`Hybrid'] Additionally, we considered a hybrid of the `Raw' and `Histogram' methods. As shown in \figureref{fig:Hybrid}, in this approach, the `Raw' and `Histogram' input forms are concatenated.

\end{description}

The architecture of the arbitration models we propose is a fully-connected layer followed by a softmax layer. We experimented with multi-layer perceptrons of different depths (from 1 layer to 4 layers), the hidden layers of different lengths (from 5 to 20), convolutional layers instead of fully connected layers, activation functions (RELU, ELU, GELU), but these parameters were found not to significantly influence performance.

For each arbitration model architecture and hyperparameter setting, we conduct five experiments on the results of each first-stage model experiment. So, when we consider the two-stage model as a whole, we run $5\times5=25$ experiments for each architecture and hyperparameter setting.

\begin{figure}[htbp]
\floatconts
  {fig:Raw and Histogram}
  {\caption{`Raw' and `Histogram' pre-processing for the arbitration model. Each small square in `Raw' is the output of the first-stage model (probability of `abnormal') for one window.
  In this example there are 16 windows in the recording. 
  In the general case, since we use the data between 1 and 21 minutes in a recording at most, a recording contains at most 20 windows with a length of 1 minute. 
  When there are fewer than 20, we pad zero at the end. Then we count the `Raw' into a histogram of ten equal bins across the range 0-10.}}
  {\includegraphics[width=0.8\linewidth]{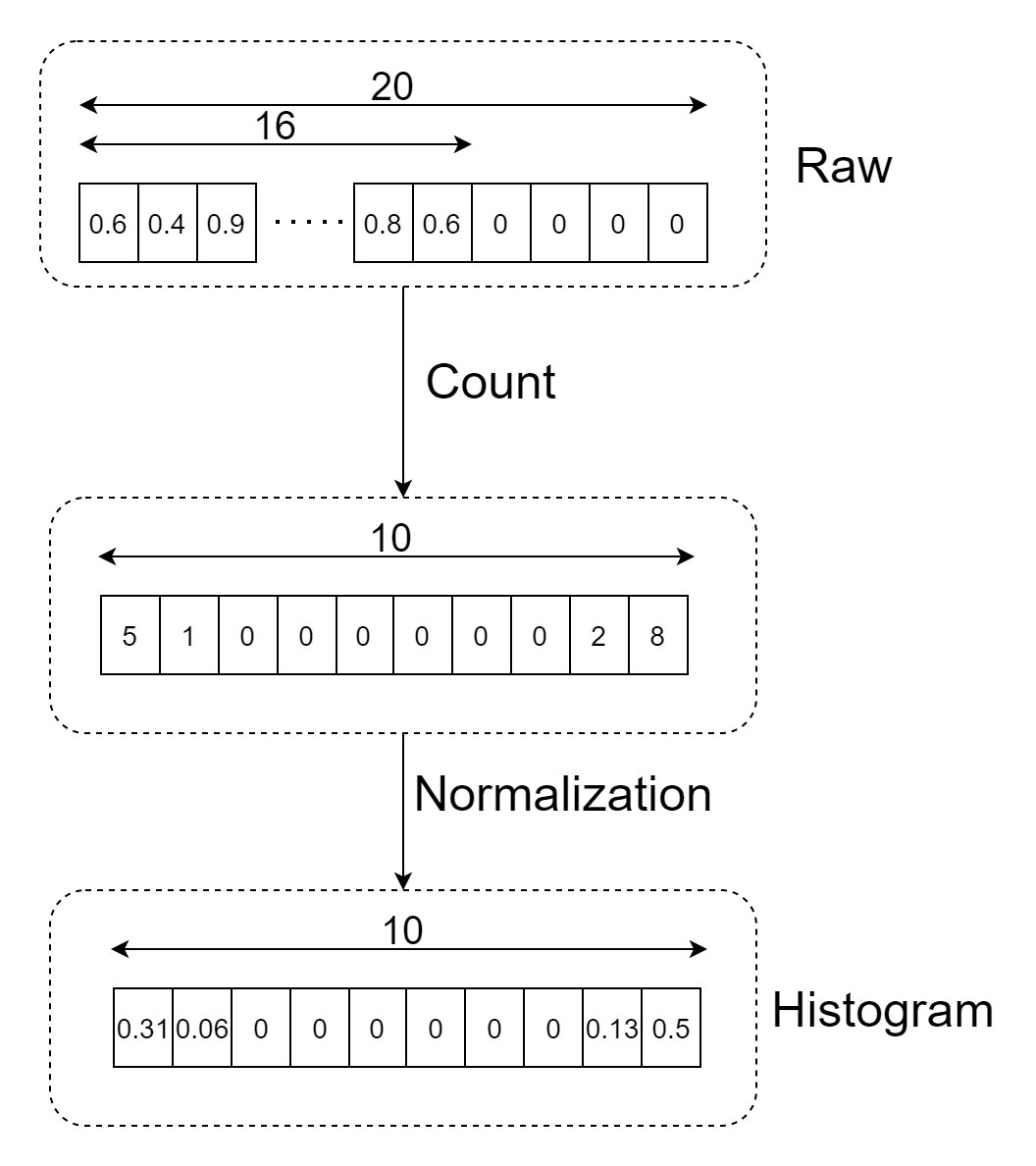}}
\end{figure}

\begin{figure}[htbp]
\floatconts
  {fig:Hybrid}
      {\caption{`Hybrid' pre-processing for the arbitration model.}}
  {\includegraphics[width=1\linewidth]{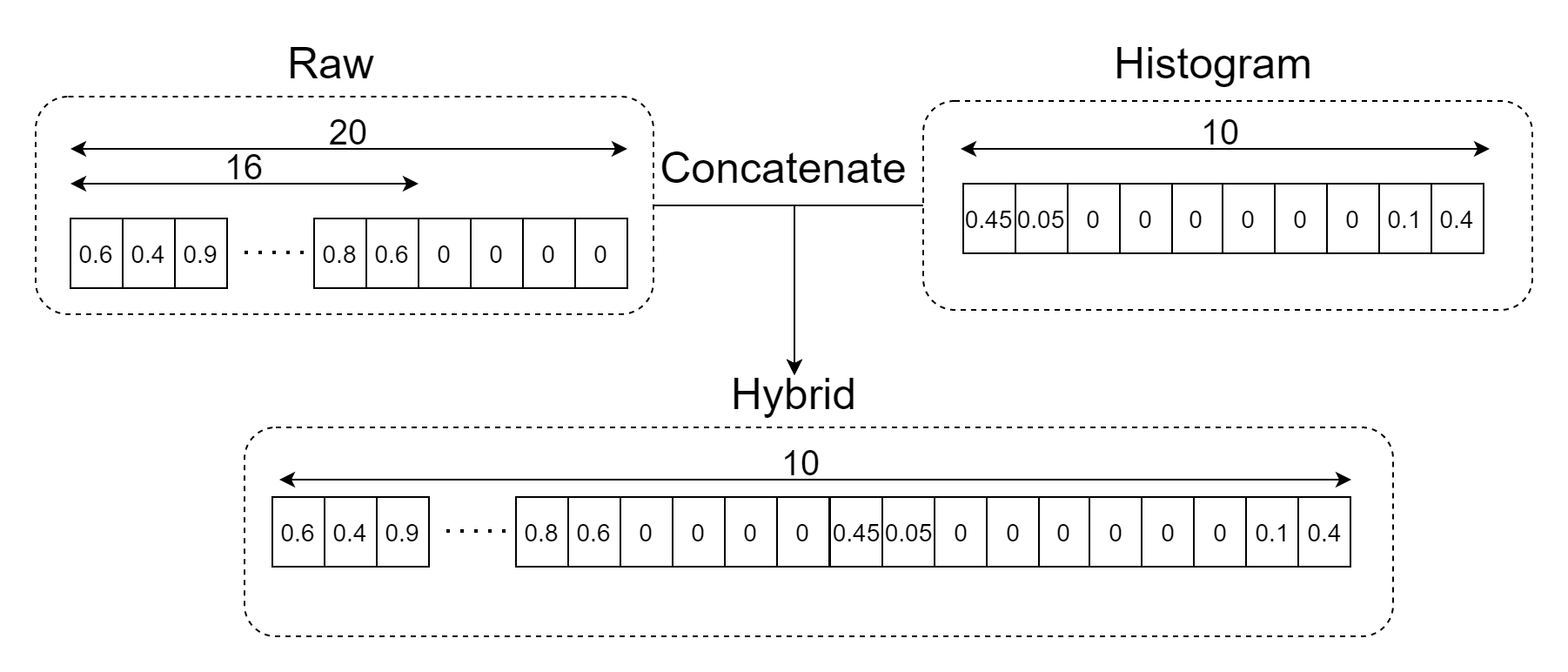}}
\end{figure}

\section{Results}
\subsection{Performance of Our Two-Stage Model}

As shown in \figureref{fig:deep4 60s and 600s,fig:window length}, all our proposed machine learning arbitration methods outperform the baseline methods (`No arbitration' and `Mean'), regardless of window length. The highest average accuracy (25 experiments) for the whole two-stage model achieved by any approach was 93.3, while the highest average accuracy for a single instance of the first-stage model (5 experiments) was 96.2, both achieved by the `Hybrid' approach with a window length of 600 s.
In addition, from \figureref{fig:window length}, we can find that the arbitration stage and increased window length both greatly improve the sensitivity of the model without substantially compromising specificity.

\begin{figure}[htbp]
\floatconts
  {fig:deep4 60s and 600s}
  {\caption{Performance of different arbitration models using window lengths of (a) 60 s and (b) 600 s. 
  Points with the same marker shape come from the same instance of the first-stage model. 
  The dashed lines represent the mean for each arbitration method.}}
  {%
    \subfigure[60 s window length]{\label{fig:60 s window length}%
      \includegraphics[width=1\linewidth]{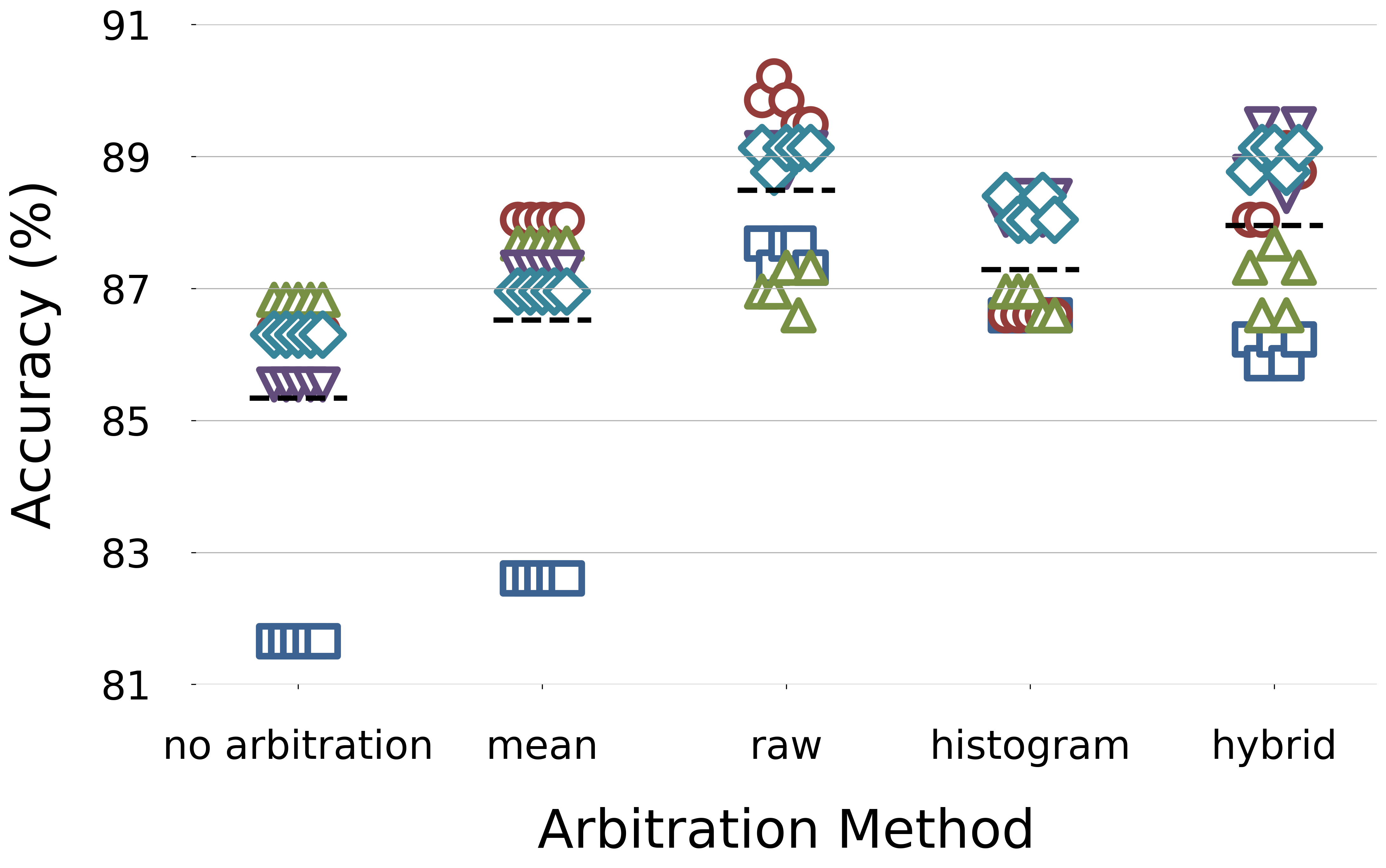}}%
    \qquad
    \subfigure[600 s window length]{\label{fig:600 s window length}%
      \includegraphics[width=1\linewidth]{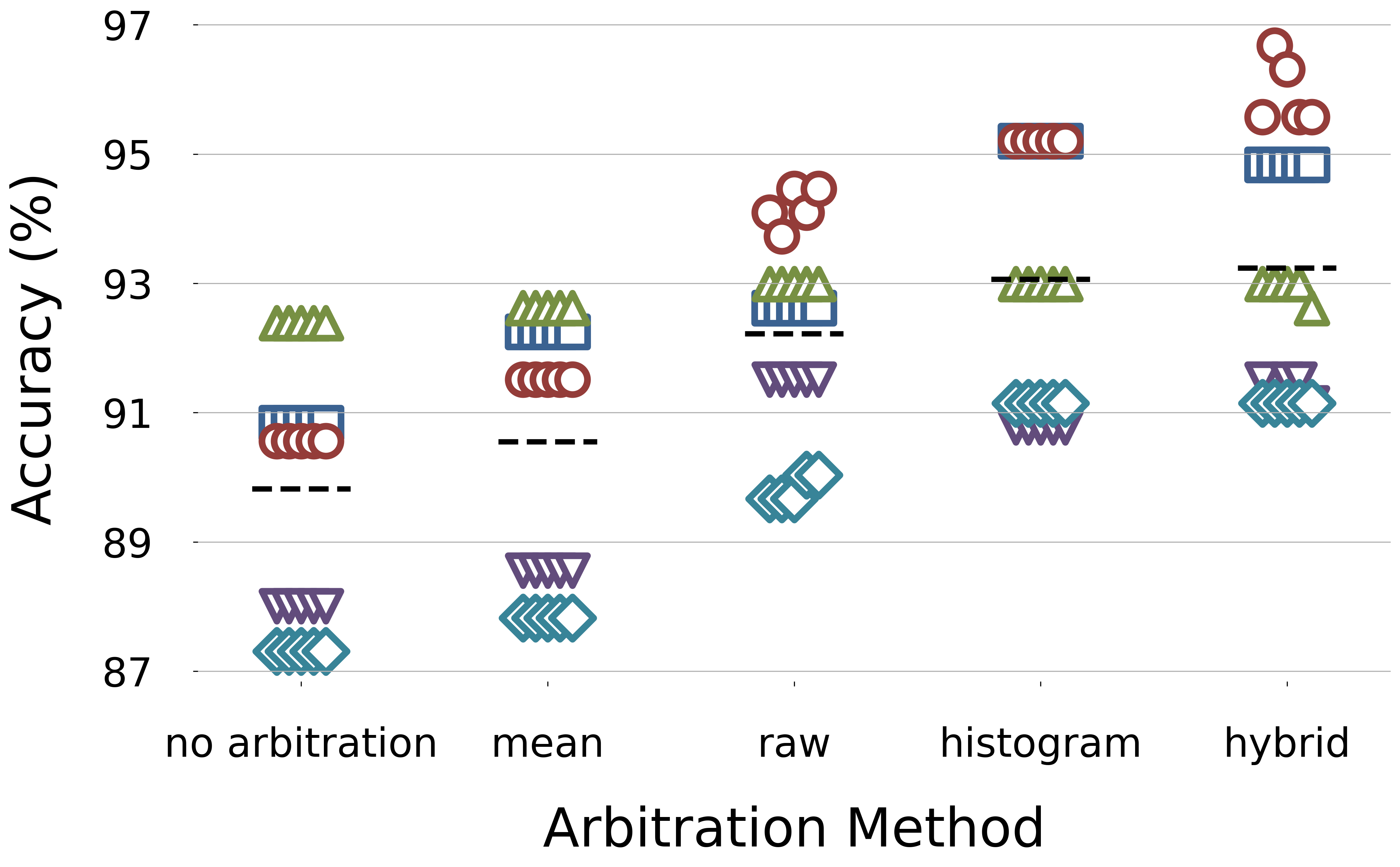}}
  }
\end{figure}

\subsection{Effect of Window Length}
As shown in \figureref{fig:window length}, both the performance of the one-stage model (`No arbitration') and the two-stage approaches gets better as the window length increases, although the performance is not strictly monotonically increasing; all models have the worst performance at 60 s and the best performance at 600 s. 
For the effect on sensitivity and specificity, we can see similar effects to the two-stage model.

\begin{figure}[htbp]
\floatconts
  {fig:window length}
  {\caption{Effect of window length on (a) accuracy, (b) sensitivity, and (c) specificity.
            Note that the accuracy of the `no\_arbitration' approach is calculated across all windows ($\num{4340} \leq N \leq \num{57482}$, depending on window length), whereas the accuracy of the arbitration models is calculated across all recordings ($N=\num{2993}$).}}
  {%
    \subfigure[accuracy]{\label{fig:accuracy}%
      \includegraphics[width=1\linewidth]{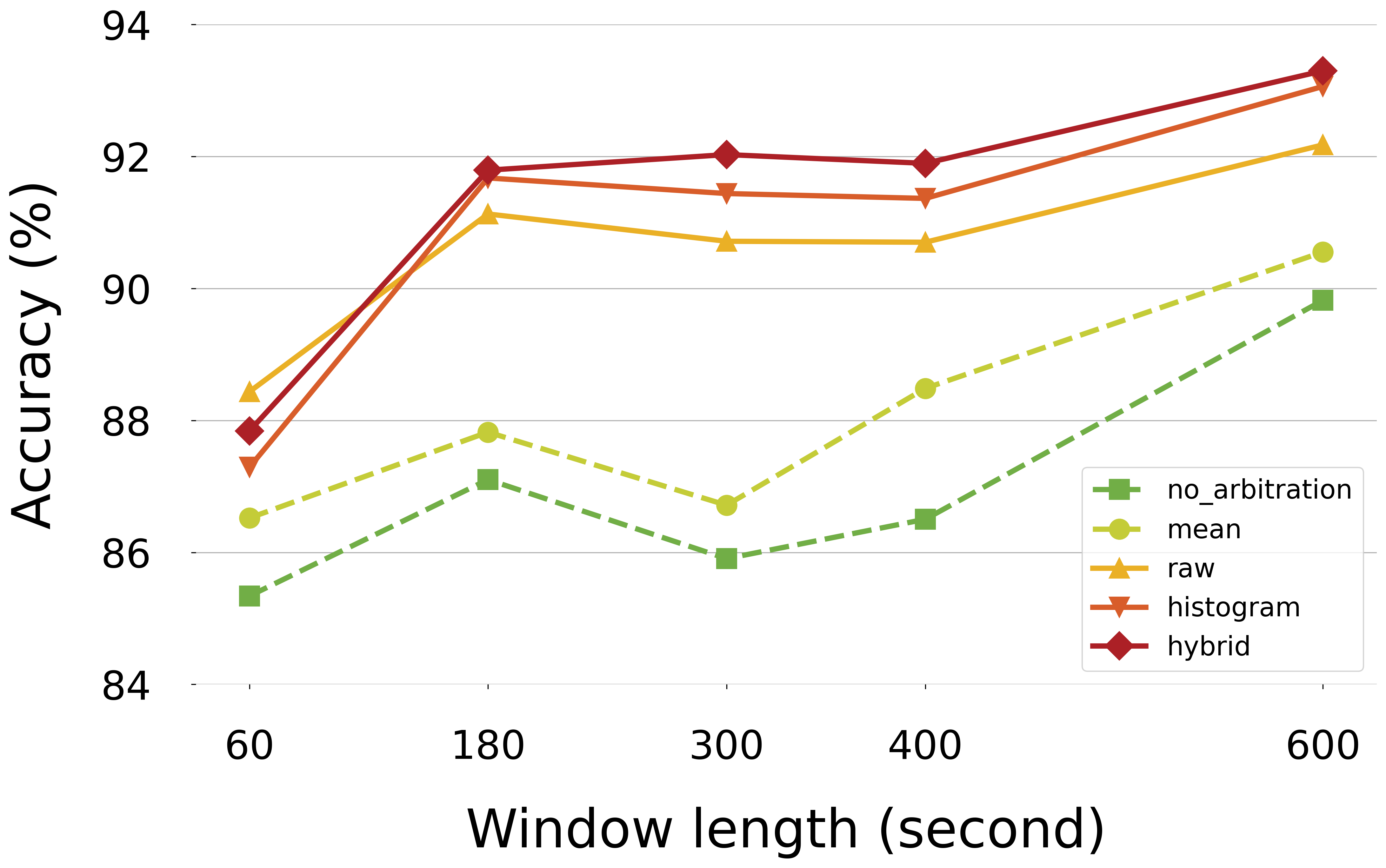}}%
    \qquad
    \subfigure[sensitivity]{\label{fig:sensitivity}%
      \includegraphics[width=1\linewidth]{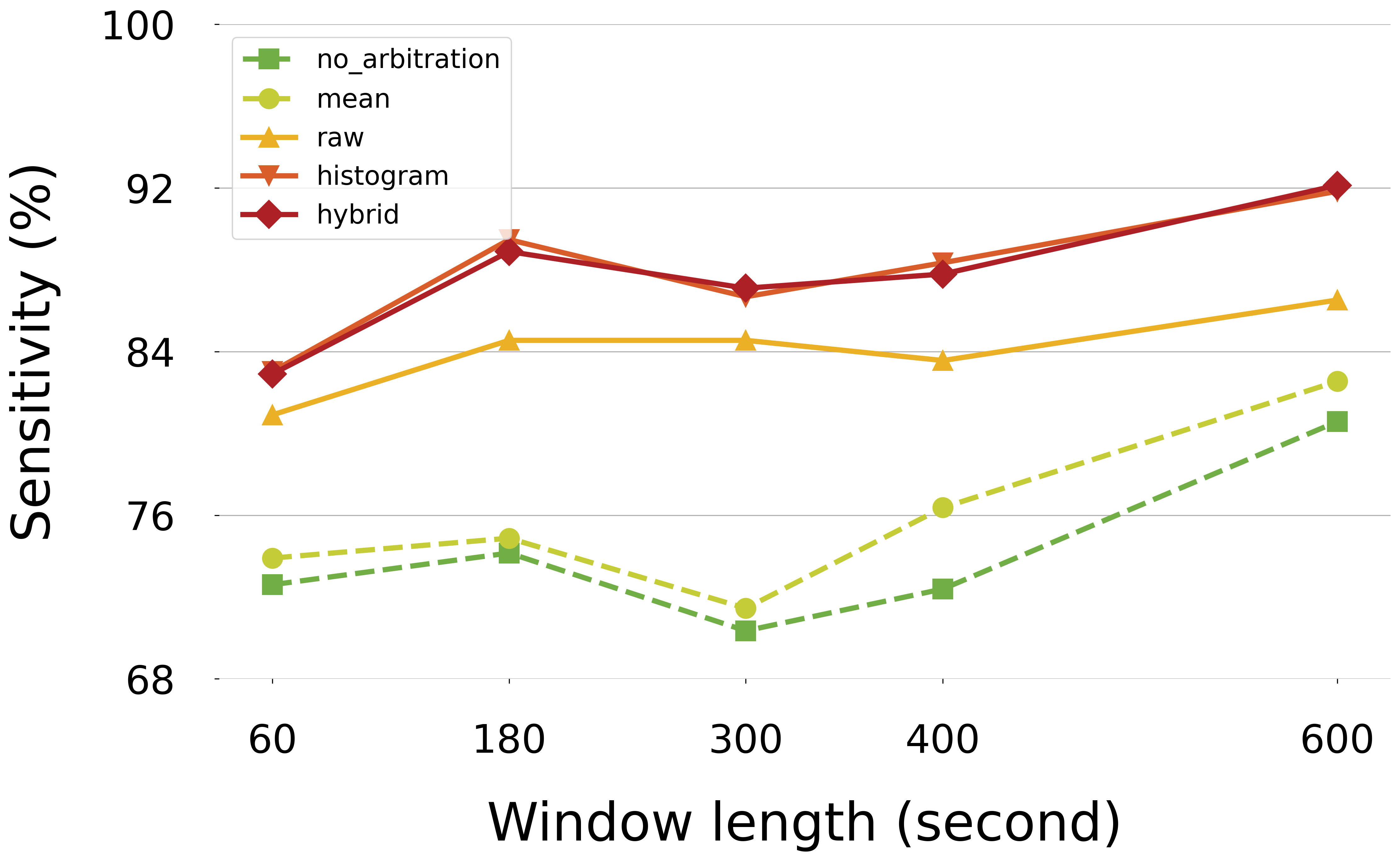}}
    \qquad
    \subfigure[specificity]{\label{fig:specificity}%
      \includegraphics[width=1\linewidth]{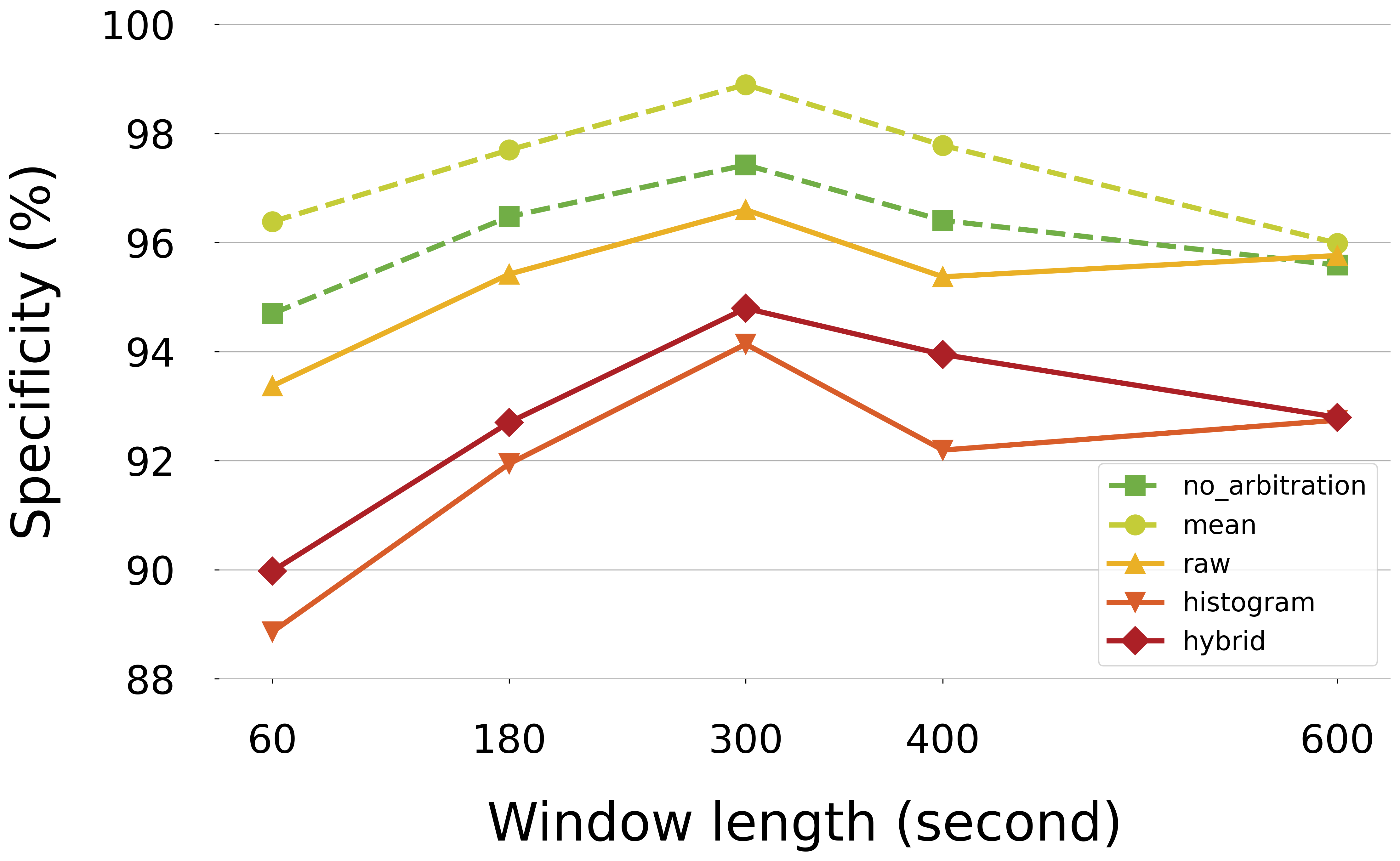}}
  }
\end{figure}

\subsection{The Search for the Arbitration Model Architecture}
As shown in \tableref{tab:hidden layer depth and length}, we examined the effect of hidden layer depth and length on the performance of the arbitration model. The results show that they have no significant impact on the model performance, although when the model depth is greater than or equal to three, the model is hard to train. (When the model parameters are initialised in a high loss position, the model will be difficult to train, that is, maintaining a high loss, although when it is initially in a relatively low loss position, the model can reach the same performance as the shallower architectures.). We also experimented with varying the activation function (RELU \citep{glorot2011deep}, ELU\citep{clevert2015fast}, GELU\citep{hendrycks2016gaussian}), but the results show that they did not affect the model performance substantially. Therefore, we finally chose to use a simple fully-connected layer and a SoftMax layer to form our model to pursue the optimisation of computational efficiency.

\begin{table}[htbp]
\floatconts
    {tab:hidden layer depth and length}
    {\caption{The effect of hidden layer depth and length on the performance of the arbitration model}}%

    \resizebox{\linewidth}{!}
    {
    \begin{tabular}{|c|c|c|c|c|}
    \hline
    \multirow{2}*{\abovestrut{2.2ex} \bfseries Hidden layer depth} & \multicolumn{4}{c|}{\abovestrut{2.2ex}\bfseries Hidden layer length}\\\cline{2-5}& \bfseries 5 & \bfseries 10 &\bfseries 15 &\bfseries 20\\\hline
    \abovestrut{2.2ex}\bfseries 0 & 0.9330 & 0.9330 & 0.9330 & 0.9330\\
    \bfseries 1 & 0.9333 & 0.9342 & 0.9334 & 0.9339 \\
    \bfseries 2 & 0.9176 & 0.9334 & 0.9328 & 0.9325 \\
    \belowstrut{0.2ex} \bfseries 3 & 0.6686 & 0.6732 & 0.6102 & 0.7045\\\hline
    \end{tabular} 
    }
   
\end{table}

\section{Discussion}

\subsection{Window Length (Scope)}

As noted in Section \ref{sec:background}, it can be argued that EEG window labels inherited from the full recording/session label are often misleading. The smaller the window, the less it represents the wider recording.
In particular, a single transient abnormal event may be sufficient for a recording to be labelled as `abnormal' even when the majority of the windows contain no discernible abnormality.
Hence we hypothesised that increasing the window length would improve training (of the first stage model) by making the window labels more accurate (\conjectureref{con:WL}). 

\figureref{{fig:window length}} shows that the performance of the one-stage model (`no\_arbitration') roughly increases with increasing window length, which confirms that hypothesis. 
For a window length of 600 s, the accuracy increases by about 5 percentage points compared with the 60 s window. 
The performance of the two-stage models show similar trends.

\subsection{The Second-Stage Model (Arbitration)}

As seen in \figureref{{fig:deep4 60s and 600s},{fig:window length}}, all our proposed variants of the arbitration model improve accuracy substantially compared with the one-stage model (`No arbitration') and baseline arbitration model (`Mean'). 
`Mean' offers minor improvement over `No arbitration': less than two percentage points.
As a simple non-parametric algorithm, `Mean' can mitigate occasional anomalous outputs but cannot learn more complex or finely tuned decisions. 
All neural network arbitration models outperformed the baseline methods, confirming \conjectureref{con:arbitration}.

Comparing panels in \figureref{fig:window length} indicates that the improvements in accuracy (both for increased window length and the proposed arbitration models) are underpinned by improved sensitivity with relatively little compromise, if any, in specificity.
This supports our supposition that, when many small windows inherit an abnormal label from their parent recording, many of the resulting labels are misleading; the window many contain no evident abnormalities, leading to increased `false negative' results.

As discussed in the Appendix, we experimented with applying the arbitration models to alternative first case architectures.
Similar benefits were observed when using a Vision Transformer (ViT), but not when using a Temporal Convolutional Network (TCN).


\figureref{fig:window length}\subfigref{fig:accuracy} shows that machine learning arbitration models outperform the baseline methods across all window lengths in our experiments, although the effect is less pronounced at a window length of 60 s.

Although evidence of arbitration stages can be found in the codebase of previous studies such as that of \citet{schirrmeister2017deep}, the concept and the selection of the model are not discussed, suggesting they were not considered to be important. 
We have proved that using a machine learning arbitration method can substantially exceed the baseline performance of the same model. 

In some other approaches, such as that of \citet{alhussein2019eeg}, arbitration of classifier outputs is not applicable because a single model fuses features across all windows to achieve a single classifier output for the recording.
This approach has sound justification, but the increased complexity of the first stage model poses a challenge for optimisation.
The results we present using relatively simple architectures demonstrate substantially greater accuracy.
This may simply reflect the ease of achieving relatively thorough optimisation for our approach.
Alternatively, the arbitration approach may present some distinct advantage in terms of robustness to transient non-clinical anomalies, which might dominate the decision in an architecture with upstream fusion of features across windows.
Confirmation of an explanation for the performance differences between these methods would require a more extensive case-by-case comparison.

\subsection{Label Quality and Performance Ceilings of Machine-Learning-Based Models on EEG binary Classification Problems}

\citet{gemein2020machine} suggested that EEG pathology decoding accuracies observed  \citet{van2019detecting,schirrmeister2017deep,roy2019chrononet} at approximately 86 percent were approaching the theoretical optimum imposed by label noise. 
This suggestion was based on the observation that inter-rater agreement in the binary classification of EEGs into pathological and non-pathological has been reported as 86–88\% \citep{houfek1959reliability,rose1973reliability}, although these scores were based on EEG ratings of only two neurologists. 
In a more recent, broader study, \citet{beuchat2021prospective} found interrater agreement to be even lower, $82-86$ percent.
However, our study demonstrated that the performance of machine learning-based models in EEG binary classification could be much greater than 86\%. 
Although different raters may give different labels to the same EEG signal, a machine learning model can learn to replicate the judgement of one rater (or team of raters, as used in the curation of TUAB \citep{lopez2017automated}). 
Now that machine learning approaches can, in some sense, match human expert performance in this task, future work should include the curation of datasets that combine a diverse range of human expert judgements and/or data on clinical outcomes to optimise label accuracy.

\subsection{Future Work}

In this study, we explored a limited range of arbitration model architectures to demonstrate the importance of arbitration in windowed EEG classifiers. 
In immediate future work, we will explore a wider range of arbitration models, such as random forests.
Arguably, the inputs to the arbitration model can be thought of as tabular data.
Random forests are frequently found to outperform neural networks on tabular data.

It is likely that our pre-processing of the first-stage outputs can also be optimised further.
We will explore the use of overlapping windows to increase the resolution of information available to the arbitration model.
Further enhancement may be achieved by optimising the binning of the `Histogram' pre-processing.
Rather than using a simple linear spacing of windows, it may be more effective to use narrower bins in ranges with a higher density of samples and wider bins (coarser resolution) elsewhere.

We will also extend the application of arbitration to cases in which the classification task spans multiple recordings from a single clinical visit, using the wider TUEG dataset in combination with automated labelling based on the text reports \citep{western2021automatic}.

In addition to efforts to improve the arbitration stage, we will continue to explore alternative first-stage architectures.
Comparing \figureref{fig:deep4 60s and 600s}\subfigref{fig:60 s window length} and \figureref{{fig:60s windows on TCN},{fig:60s windows on ViT}} indicates the degree of improvement achieved by arbitration varies significantly between different first-stage architectures.
It is possible that the best first-stage architecture for use with arbitration is not the same as the best single-stage architecture (for per-window classification, i.e. `No arbitration').
Furthermore, as we move on from TUAB to the larger TUEG dataset, we may find that data-hungry architectures such as transformers may outperform those that have achieved previous state-of-the-art results on TUAB.

The arbitration principle is likely to be transferable to other time-series applications where a holistic classification is to be applied to a windowed signal.
For example, in ECG arrhythmia detection, end-to-end training of architectures with densely connected output layers is common \citep{ebrahimi2020review}, but we are not aware of other cases where this final classification layer is trained separately.
Our results suggest that this approach is an effective way to increase the input scope of the system with minimal added computational expense.
For cardiac electrophysiology, enabling the application of machine learning classifiers to holistic analysis of long-term Holter recordings could be important for the detection of subtle abnormalities that cannot be discerned from shorter signals.

\section{Conclusion}

Our proposed approach, combining increased window length and a machine learning arbitration stage, substantially improved upon previous state-of-the-art performance in clinical EEG classification. 
The results support our premise that the inheritance of window labels from recording labels compromised the sensitivity of previous state-of-the-art solutions.
Given the importance of sensitivity for promising applications such as routine screening or accelerating the workflow of human EEG interpreters, this improvement presented here is an important step towards the broader translation of machine learning EEG classifiers into clinical practice.
The principles may also be transferable to other time-series classification problems.

\acks{This work was supported by a PhD studentship funded by Southmead Hospital Charity and the University of the West of England.}

\bibliography{jmlr-sample}

\begin{thebibliography}{23}
\providecommand{\natexlab}[1]{#1}
\providecommand{\url}[1]{\texttt{#1}}
\expandafter\ifx\csname urlstyle\endcsname\relax
  \providecommand{\doi}[1]{doi: #1}\else
  \providecommand{\doi}{doi: \begingroup \urlstyle{rm}\Url}\fi

\bibitem[Alhussein et~al.(2019)Alhussein, Muhammad, and
  Hossain]{alhussein2019eeg}
Musaed Alhussein, Ghulam Muhammad, and M~Shamim Hossain.
\newblock Eeg pathology detection based on deep learning.
\newblock \emph{IEEE Access}, 7:\penalty0 27781--27788, 2019.

\bibitem[Amin et~al.(2019)Amin, Hossain, Muhammad, Alhussein, and
  Rahman]{amin2019cognitive}
Syed~Umar Amin, M~Shamim Hossain, Ghulam Muhammad, Musaed Alhussein, and
  Md~Abdur Rahman.
\newblock Cognitive smart healthcare for pathology detection and monitoring.
\newblock \emph{IEEE Access}, 7:\penalty0 10745--10753, 2019.

\bibitem[{Bai} et~al.(2018){Bai}, Kolter, and Koltun]{bai2018empirical}
Shaojie {Bai}, J.~Zico Kolter, and Vladlen Koltun.
\newblock An empirical evaluation of generic convolutional and recurrent
  networks for sequence modeling.
\newblock 2018.
\newblock \doi{10.48550/ARXIV.1803.01271}.
\newblock URL \url{https://arxiv.org/abs/1803.01271}.

\bibitem[Banville et~al.(2021)Banville, Chehab, Hyv{\"a}rinen, Engemann, and
  Gramfort]{banville2021uncovering}
Hubert Banville, Omar Chehab, Aapo Hyv{\"a}rinen, Denis-Alexander Engemann, and
  Alexandre Gramfort.
\newblock Uncovering the structure of clinical eeg signals with self-supervised
  learning.
\newblock \emph{Journal of Neural Engineering}, 18\penalty0 (4):\penalty0
  046020, 2021.

\bibitem[Banville et~al.(2022)Banville, Wood, Aimone, Engemann, and
  Gramfort]{banville2022robust}
Hubert Banville, Sean~UN Wood, Chris Aimone, Denis-Alexander Engemann, and
  Alexandre Gramfort.
\newblock Robust learning from corrupted eeg with dynamic spatial filtering.
\newblock \emph{NeuroImage}, 251:\penalty0 118994, 2022.

\bibitem[Beuchat et~al.(2021)Beuchat, Alloussi, Reif, Sterlepper, Rosenow, and
  Strzelczyk]{beuchat2021prospective}
Isabelle Beuchat, Senubia Alloussi, Philipp~S Reif, Nora Sterlepper, Felix
  Rosenow, and Adam Strzelczyk.
\newblock Prospective evaluation of interrater agreement between eeg
  technologists and neurophysiologists.
\newblock \emph{Scientific Reports}, 11\penalty0 (1):\penalty0 13406, 2021.

\bibitem[Clevert et~al.(2015)Clevert, Unterthiner, and
  Hochreiter]{clevert2015fast}
Djork-Arn{\'e} Clevert, Thomas Unterthiner, and Sepp Hochreiter.
\newblock Fast and accurate deep network learning by exponential linear units
  (elus).
\newblock \emph{arXiv preprint arXiv:1511.07289}, 2015.

\bibitem[Dosovitskiy et~al.(2020)Dosovitskiy, Beyer, Kolesnikov, Weissenborn,
  Zhai, Unterthiner, Dehghani, Minderer, Heigold, Gelly,
  et~al.]{dosovitskiy2020image}
Alexey Dosovitskiy, Lucas Beyer, Alexander Kolesnikov, Dirk Weissenborn,
  Xiaohua Zhai, Thomas Unterthiner, Mostafa Dehghani, Matthias Minderer, Georg
  Heigold, Sylvain Gelly, et~al.
\newblock An image is worth 16x16 words: Transformers for image recognition at
  scale.
\newblock \emph{arXiv preprint arXiv:2010.11929}, 2020.

\bibitem[Ebrahimi et~al.(2020)Ebrahimi, Loni, Daneshtalab, and
  Gharehbaghi]{ebrahimi2020review}
Zahra Ebrahimi, Mohammad Loni, Masoud Daneshtalab, and Arash Gharehbaghi.
\newblock A review on deep learning methods for {ECG} arrhythmia
  classification.
\newblock \emph{Expert Systems with Applications: X}, 7:\penalty0 100033,
  September 2020.
\newblock ISSN 2590-1885.
\newblock \doi{10.1016/j.eswax.2020.100033}.
\newblock URL
  \url{https://www.sciencedirect.com/science/article/pii/S2590188520300123}.

\bibitem[Gemein et~al.(2020)Gemein, Schirrmeister, Chrab{\k{a}}szcz, Wilson,
  Boedecker, Schulze-Bonhage, Hutter, and Ball]{gemein2020machine}
Lukas~AW Gemein, Robin~T Schirrmeister, Patryk Chrab{\k{a}}szcz, Daniel Wilson,
  Joschka Boedecker, Andreas Schulze-Bonhage, Frank Hutter, and Tonio Ball.
\newblock Machine-learning-based diagnostics of eeg pathology.
\newblock \emph{NeuroImage}, 220:\penalty0 117021, 2020.

\bibitem[Glorot et~al.(2011)Glorot, Bordes, and Bengio]{glorot2011deep}
Xavier Glorot, Antoine Bordes, and Yoshua Bengio.
\newblock Deep sparse rectifier neural networks.
\newblock In \emph{Proceedings of the fourteenth international conference on
  artificial intelligence and statistics}, pages 315--323. JMLR Workshop and
  Conference Proceedings, 2011.

\bibitem[Hendrycks and Gimpel(2016)]{hendrycks2016gaussian}
Dan Hendrycks and Kevin Gimpel.
\newblock Gaussian error linear units (gelus).
\newblock \emph{arXiv preprint arXiv:1606.08415}, 2016.

\bibitem[Houfek and Ellingson(1959)]{houfek1959reliability}
Edward~E Houfek and Robert~J Ellingson.
\newblock On the reliability of clinical eeg interpretation.
\newblock \emph{The Journal of nervous and mental disease}, 128\penalty0
  (5):\penalty0 425--437, 1959.

\bibitem[L{\'o}pez et~al.(2017)L{\'o}pez, Obeid, and
  Picone]{lopez2017automated}
Silvia L{\'o}pez, I~Obeid, and J~Picone.
\newblock \emph{Automated interpretation of abnormal adult
  electroencephalograms}.
\newblock PhD thesis, 2017.

\bibitem[Muhammad et~al.(2020)Muhammad, Hossain, and Kumar]{muhammad2020eeg}
Ghulam Muhammad, M~Shamim Hossain, and Neeraj Kumar.
\newblock Eeg-based pathology detection for home health monitoring.
\newblock \emph{IEEE Journal on Selected Areas in Communications}, 39\penalty0
  (2):\penalty0 603--610, 2020.

\bibitem[Obeid and Picone(2016)]{obeid2016temple}
Iyad Obeid and Joseph Picone.
\newblock The temple university hospital eeg data corpus.
\newblock \emph{Frontiers in neuroscience}, 10:\penalty0 196, 2016.

\bibitem[Rose et~al.(1973)Rose, Penry, White, and Sato]{rose1973reliability}
Stephen~W Rose, J~Kiffin Penry, Billy~G White, and Susumu Sato.
\newblock Reliability and validity of visual eeg assessment in third grade
  children.
\newblock \emph{Clinical Electroencephalography}, 4\penalty0 (4):\penalty0
  197--205, 1973.

\bibitem[Roy et~al.(2019)Roy, Kiral-Kornek, and Harrer]{roy2019chrononet}
Subhrajit Roy, Isabell Kiral-Kornek, and Stefan Harrer.
\newblock Chrononet: a deep recurrent neural network for abnormal eeg
  identification.
\newblock In \emph{Artificial Intelligence in Medicine: 17th Conference on
  Artificial Intelligence in Medicine, AIME 2019, Poznan, Poland, June 26--29,
  2019, Proceedings 17}, pages 47--56. Springer, 2019.

\bibitem[Schirrmeister et~al.(2017)Schirrmeister, Springenberg, Fiederer,
  Glasstetter, Eggensperger, Tangermann, Hutter, Burgard, and
  Ball]{schirrmeister2017deep}
Robin~Tibor Schirrmeister, Jost~Tobias Springenberg, Lukas Dominique~Josef
  Fiederer, Martin Glasstetter, Katharina Eggensperger, Michael Tangermann,
  Frank Hutter, Wolfram Burgard, and Tonio Ball.
\newblock Deep learning with convolutional neural networks for eeg decoding and
  visualization.
\newblock \emph{Human brain mapping}, 38\penalty0 (11):\penalty0 5391--5420,
  2017.

\bibitem[Van~Leeuwen et~al.(2019)Van~Leeuwen, Sun, Tabaeizadeh, Struck,
  Van~Putten, and Westover]{van2019detecting}
KG~Van~Leeuwen, H~Sun, M~Tabaeizadeh, AF~Struck, MJAM Van~Putten, and
  MB~Westover.
\newblock Detecting abnormal electroencephalograms using deep convolutional
  networks.
\newblock \emph{Clinical neurophysiology}, 130\penalty0 (1):\penalty0 77--84,
  2019.

\bibitem[Wagh and Varatharajah(2020)]{wagh2020eeg}
Neeraj Wagh and Yogatheesan Varatharajah.
\newblock Eeg-gcnn: Augmenting electroencephalogram-based neurological disease
  diagnosis using a domain-guided graph convolutional neural network.
\newblock In \emph{Machine Learning for Health}, pages 367--378. PMLR, 2020.

\bibitem[Western et~al.(2021)Western, Weber, Kandasamy, May, Taylor, Zhu, and
  Canham]{western2021automatic}
D~Western, T~Weber, R~Kandasamy, F~May, S~Taylor, Y~Zhu, and L~Canham.
\newblock Automatic report-based labelling of clinical eegs for classifier
  training.
\newblock In \emph{2021 IEEE Signal Processing in Medicine and Biology
  Symposium (SPMB)}, pages 1--6. IEEE, 2021.

\bibitem[Y{\i}ld{\i}r{\i}m et~al.(2020)Y{\i}ld{\i}r{\i}m, Baloglu, and
  Acharya]{yildirim2020deep}
{\"O}zal Y{\i}ld{\i}r{\i}m, Ulas~Baran Baloglu, and U~Rajendra Acharya.
\newblock A deep convolutional neural network model for automated
  identification of abnormal eeg signals.
\newblock \emph{Neural Computing and Applications}, 32:\penalty0 15857--15868,
  2020.

\end{thebibliography}

\appendix

\section{Results on Alternative First-Stage Architectures}\label{apd:first}

As shown in \figureref{fig:TCN and ViT}, we also explored the effect of the arbitration models on two alternative first-stage architectures: a temporal convolutional networks (TCN) \citep{bai2018empirical, gemein2020machine} and vision transformer (ViT) \citep{dosovitskiy2020image}. 
Full details of the implementation and hyperparameter tuning are beyond the scope of this paper, but the implementations are available in our code repository. 
We present the results briefly here to demonstrate the extent to which our method is transferrable to other first-stage models.

For TCN, the performance of the proposed arbitration models is not substantially different from the baseline (`Mean'). 
For ViT, our proposed arbitration models can provide about two percentage points of performance improvement.
Based on the present evidence, the proposed methods appear to offer a safe improvement in the sense that no cases were observed in which accuracy was substantially worsened.
We will test the effect of the arbitration models on a wider selection of first-stage models as well as longer window lengths in future work.

\begin{figure}[htbp]
\floatconts
  {fig:TCN and ViT}
  {\caption{Performance of different arbitration methods using (a) TCN and (b) ViT as the first-stage architecture with a window length of 60 s.}}
  {%
    \subfigure[TCN]{\label{fig:60s windows on TCN}%
      \includegraphics[width=1\linewidth]{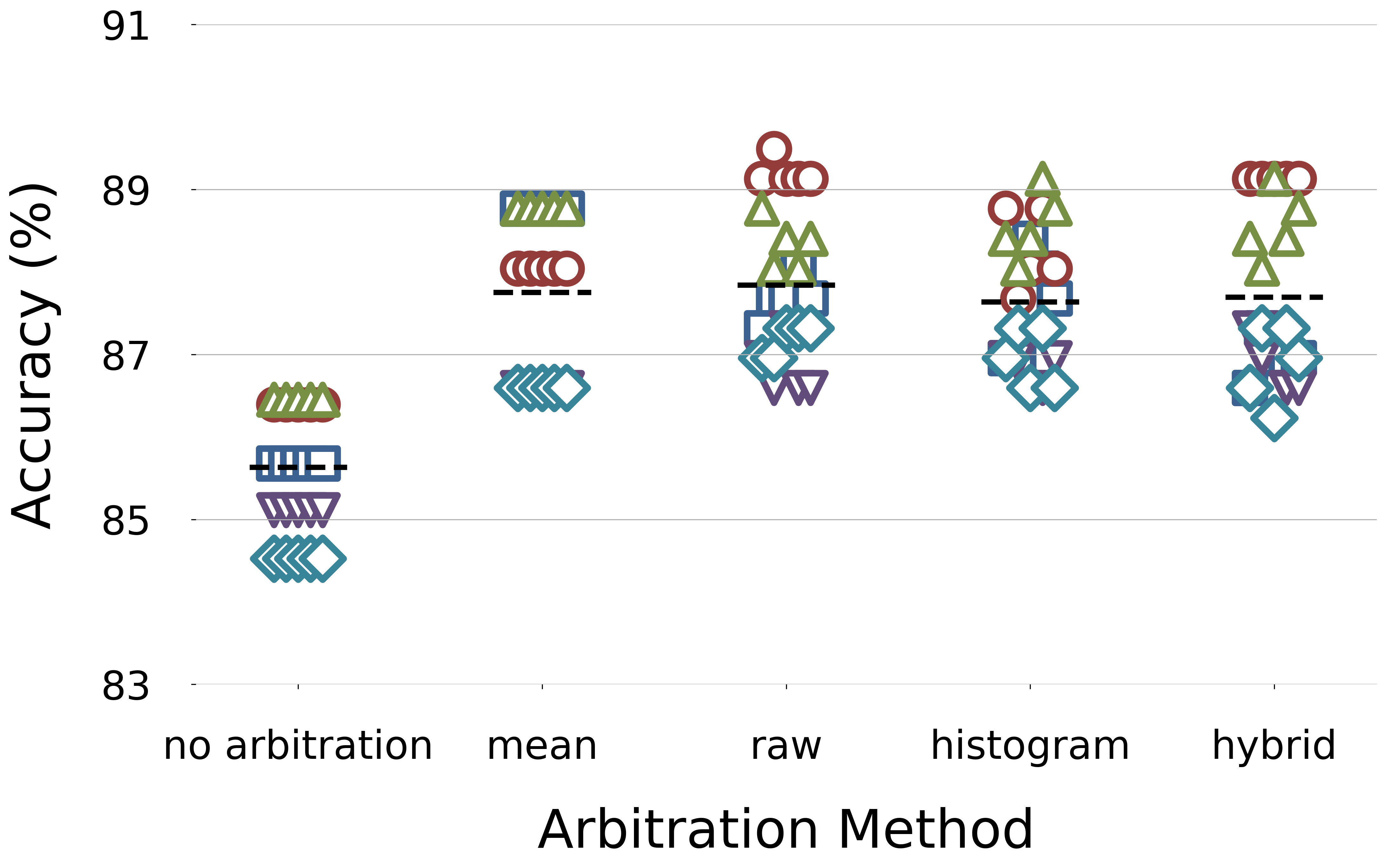}}%
    \qquad
    \subfigure[ViT]{\label{fig:60s windows on ViT}%
      \includegraphics[width=1\linewidth]{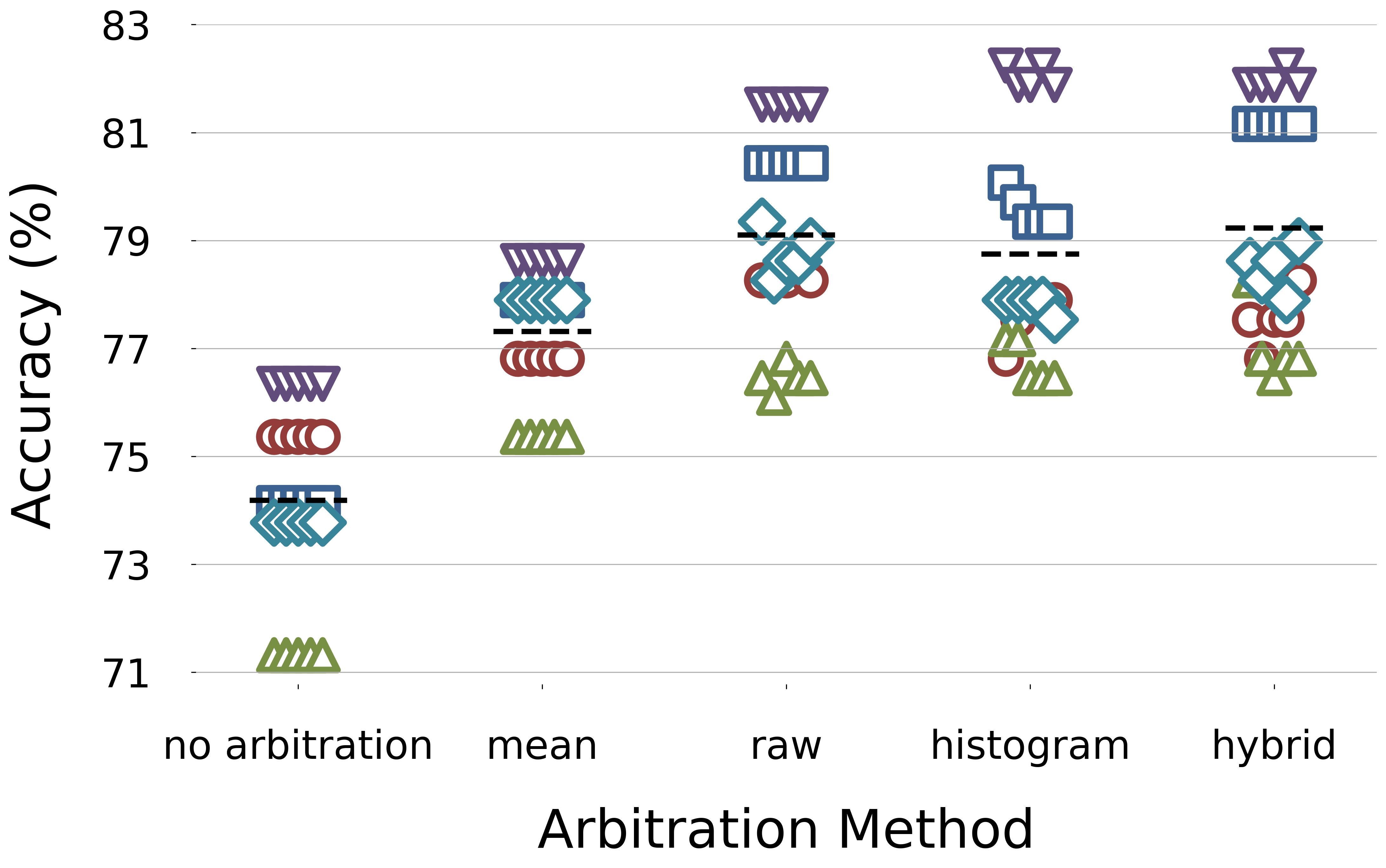}}
  }
\end{figure}

%

\end{document}